\documentclass[11pt,a4paper]{article}
\usepackage[hyperref]{acl2021}
\usepackage{times}
\usepackage{latexsym}

\usepackage{microtype}

\aclfinalcopy % Uncomment this line for the final submission
\def\aclpaperid{143} %  Enter the acl Paper ID here

\usepackage{color}
\usepackage{colortbl}
\usepackage{tabularx}
\usepackage{latexsym}
\usepackage{amsmath}
\usepackage{amssymb}
\usepackage{multirow}
\usepackage{here}
\usepackage{setspace}
\usepackage{tikz,tikz-qtree}
\usetikzlibrary{shapes.geometric,positioning,shapes,fit,arrows,backgrounds}
\usetikzlibrary{intersections, calc}
\usepackage{algorithm}
\usepackage{algorithmic}
\usepackage{drs}
\let\drsalignment=l
\usepackage{proof}
\usepackage{ebproof}
\newcommand{\todo}[1]{\textcolor{black}{#1}}
\newcommand{\pending}[1]{\textcolor{black}{#1}}
\usepackage{textcomp}
\usepackage[T1]{fontenc}
\usepackage{ulem}
\usepackage{url}
\usepackage{mediabb}
\usepackage[subrefformat=parens]{subcaption}
\usepackage{gb4ea}
\newcommand{\sem}[1]{[\![ #1 ]\!]}

\newcommand{\catS}{\textsc{S}}
\newcommand{\catPN}{\textsc{PN}}
\newcommand{\catNP}{\textsc{NP}}
\newcommand{\catVP}{\textsc{VP}}

\newcommand{\catIV}{\textsc{IV}}
\newcommand{\catIVdash}{\textsc{IV}$'$}
\newcommand{\catTV}{\textsc{TV}}
\newcommand{\catQ}{\textsc{Q}}
\newcommand{\catN}{\textsc{N}}

\newcommand{\catAdj}{\textsc{Adj}}
\newcommand{\catAdv}{\textsc{Adv}}

\newcommand{\catSbar}{\ensuremath{\overline{\mbox{\textsc{S}}}}}

\newcommand{\LF}[1]{\ensuremath{\mathbf{#1}}}
\newcommand{\drgvar}[1]{\texttt{\textbf{#1}}}
\newcommand{\righte}{$G\Rightarrow P$}
\newcommand{\lefte}{$G\Leftarrow P$}
\newcommand{\bie}{$G\Leftrightarrow P$}
\newcommand{\modelcite}[1]{\citeauthor{#1},\ \citeyear{#1}}

\newcommand{\Repl}[1]{\textsc{#1}}

\definecolor{applegreen}{rgb}{0.55, 0.71, 0.0}
\newcommand{\ModifierColor}[1]{\redbox{#1}}
\newcommand{\QuantifierColor}[1]{\bluebox{#1}}
\newcommand{\NegationColor}[1]{\greenbox{#1}}

\definecolor{olive} {cmyk} {0.64, 0.00, 0.95, 0.40}
\newcommand{\corr}[1]{\hspace{3em}\textcolor{olive}{#1}}
\newcommand{\wrong}[1]{\hspace{3em}\textcolor{red}{#1}}

\newcommand{\pospol}[1]{#1\ensuremath{^{\textcolor{red}{\uparrow}}}}
\newcommand{\negpol}[1]{#1\ensuremath{^{\textcolor{blue}{\downarrow}}}}

\usepackage{tcolorbox}
\newtcbox{\redbox}{on line,
  colframe=red,colback=red!5!white,
  boxrule=0.3mm,
  arc=0.5mm,
  boxsep=1pt,left=1pt,right=1pt,top=1pt,bottom=1pt
  }

\newtcbox{\bluebox}{on line,
  colframe=blue,colback=blue!5!white,
  boxrule=0.3mm,
  arc=0.5mm,
  boxsep=1pt,left=1pt,right=1pt,top=1pt,bottom=1pt
  }

\newtcbox{\greenbox}{on line,
  colframe=applegreen,colback=applegreen!5!white,
  boxrule=0.3mm,
  arc=0.5mm,
  boxsep=1pt,left=1pt,right=1pt,top=1pt,bottom=1pt
  }
  
\newtcbox{\orangebox}{on line,
  colframe=orange,colback=orange!10!white,
  boxrule=0.3mm,
  arc=1.0mm,
  boxsep=0pt,left=1pt,right=1pt,top=1pt,bottom=1pt
  }

\newtcbox{\blueboxa}{on line,
  colframe=blue!70!black,colback=blue!5!white,
  boxrule=0.3mm,
  arc=1.0mm,
  boxsep=0pt,left=1.2pt,right=1.2pt,top=1.3pt,bottom=1.3pt
  }

\newtcbox{\blueboxb}{on line,
  colframe=blue!70!black,colback=blue!5!white,
  boxrule=0.3mm,
  arc=1.0mm,
  boxsep=0pt,left=1.2pt,right=1.2pt,top=1.6pt,bottom=1.6pt
  }

\newtcbox{\graybox}{on line,
  colframe=white,colback=gray!15!white,
  boxrule=0mm,left=1pt,right=1pt,top=1pt,bottom=1pt
  }
\title{SyGNS: A Systematic Generalization Testbed\\Based on Natural Language Semantics}

\author{
  \parbox{\linewidth}{\centering
   Hitomi Yanaka$^{1,2}$,
   Koji Mineshima$^3$, and
   Kentaro Inui$^{4,2}$
  }
  \\
   $^1$\mbox{\rm The University of Tokyo,}
   $^2$\mbox{\rm RIKEN,}
   $^3$\mbox{\rm Keio University,}
   $^4$\mbox{\rm Tohoku University} 
  \\
  \parbox{\linewidth}{\centering
   {\tt hyanaka@is.s.u-tokyo.ac.jp},
   {\tt minesima@abelard.flet.keio.ac.jp},
   {\tt inui@ecei.tohoku.ac.jp}
   }
}

\date{}

\begin{document}
\maketitle
\begin{abstract}
Recently, deep neural networks (DNNs) have achieved great success in semantically challenging NLP tasks, yet it remains unclear whether DNN models can capture \textit{compositional} meanings, those aspects of meaning that have been long studied in formal semantics.
To investigate this issue, we propose a Systematic Generalization testbed based on Natural language Semantics (SyGNS), whose challenge is to map natural language sentences to multiple forms of scoped meaning representations, designed to account for various semantic phenomena.
Using SyGNS, we test whether neural networks can systematically parse sentences involving novel combinations of logical expressions such as quantifiers and negation.
\todo{Experiments show that Transformer and GRU models can generalize to unseen combinations of quantifiers, negations, and modifiers that are similar to given training instances in form, but not to the others.
We also find that the generalization performance to unseen combinations is better when the form of meaning representations is simpler.}
The data and code for SyGNS are publicly available at \url{https://github.com/verypluming/SyGNS}.
\end{abstract}

\section{Introduction}
\label{sec:intro}
Deep neural networks (DNNs) have shown impressive performance in various language understanding tasks~\citep[][i.a.]{NIPS2019_8589,wang2018glue}, including semantically challenging tasks such as Natural Language Inference (NLI;~\modelcite{series/synthesis/2013Dagan};~\modelcite{Bowman2015}).
However, a number of studies to probe DNN models with various NLI datasets~\cite{naik-etal-2018-stress,dasgupta2018evaluating,yanaka-etal-2019-neural,kim-etal-2019-probing,Richardson2019,saha-etal-2020-conjnli,geiger-etal-2020-neural} 
have reported that current DNN models have some limitations to generalize to diverse semantic phenomena, 
and it is still not clear whether DNN models obtain the ability to capture \textit{compositional} aspects of meaning in natural language.

There are two issues to consider here.
First, recent analyses~\cite{talmor-berant-2019-multiqa,liu-etal-2019-inoculation,mccoy-etal-2019-right} have pointed out that
the standard paradigm for evaluation, where a test set is drawn from the same distribution as the training set,
does not always indicate
that the model has obtained the intended generalization ability for language understanding.
Second, the NLI task of predicting the relationship between a premise sentence and an associated hypothesis without asking their semantic interpretation tends to be \textit{black-box}, in that it is often difficult to isolate the reasons why models make incorrect predictions~\cite{bos-2008-lets}.

%Figure 1
\begin{figure}[tb]
\centering
\scalebox{0.67}{%\displaystyle
\begin{tikzpicture}
\tikzset{headline1/.style={text width=6.0cm}};
\tikzset{sentence1/.style={text width=5.5cm}};
\tikzset{sentence2/.style={text width=6.5cm}};

%\draw[help lines] (-5,0) grid (5,10); % for debug
%\node (origin) at (0,0) {\textbullet}; % for debug
%\draw[thick,densely dashed] (-8.0,6.5) -- (3.4,6.5);

% Training
\node[headline1] (train) at (-4,10.0) {\graybox{\textbf{Training Sentences}}};

\node[sentence1] (train1) at (-4,9.2) {One \ModifierColor{wild} dog ran};
\node[sentence1] (train2) at (-4,8.4) {\QuantifierColor{All} dogs ran};
\node[sentence1] (train3) at (-4,7.6) {One dog \NegationColor{did not} run};

% Generalization Test
\node[headline1] (test) at (1.8,10.0) {\graybox{\textbf{Generalization Test}}};

\node[sentence1] (test1) at (1.8,9.1) {\QuantifierColor{All} \ModifierColor{wild} dogs ran};
\node[sentence1] (test2) at (1.8,8.0) {\QuantifierColor{All} dogs \NegationColor{did not} run};

\node[sentence1] at (-0.5,9.3) {\textsc{modifier}};
\draw[dashed,thick,red,rounded corners=1pt] (-5.5,9.4) -- (-5.5,9.5) -- (0.3,9.5) -- (0.3,9.3);

\node[sentence1] at (-0.5,8.5) {\textsc{quantifier}};
\draw[dashed,thick,blue,rounded corners=1pt] 
(-6.4,8.6) -- (-6.4,8.7) [rounded corners=5pt] -- (-0.6,8.7) -- (-0.6,8.9);
\draw[dashed,thick,blue,rounded corners=1pt]
 (-6.4,8.6) -- (-6.4,8.7) [rounded corners=5pt] -- (-0.6,8.7)  -- (-0.6,8.2);

\node[sentence1] at (-0.5,7.0) {\textsc{negation}};
\draw[dashed,thick,applegreen,rounded corners=1pt] (-4.5,7.4) -- (-4.5,7.2) -- (1.3,7.2) -- (1.3,7.8);

% Meaning Representations
\node[headline1] at (-4,6.2) {\graybox{\textbf{Multiple meaning representations}}};
\node[sentence2] at (-4,5.4)
{\small MR1: $\forall x.(\negpol{\LF{dog}}(x) \wedge \negpol{\LF{wild}}(x))
 \to (\pospol{\LF{run}}(x))$};
\node[sentence2] at (-4,4.8) {\small MR2: \texttt{ALL AND DOG WILD RUN}};
\node[sentence2] at (-4,3.6) {\small MR3: \drs{}{\ifdrs{$x_1$}{$\LF{wild}(x_1)$\\$\LF{dog}(x_1)$}{}{$\LF{run}(x_1)$}}};

% Evaluation
\node[headline1] at (2.5,6.2) {\graybox{\textbf{Evaluation methods}}};
\node[sentence1] at (2.5,5.6)
 {\small \underline{Exact matching}: $G=P$?};
\node[sentence1] at (2.5,5.1)
 {\small \underline{Theorem Proving}: $G \Leftrightarrow P$?};
\node[sentence1] at (2.5,4.6)
 {\small \underline{Polarity}: $\{\negpol{\LF{dog}},\negpol{\LF{wild}}, \pospol{\LF{run}}\}$};
\node[sentence1] at (2.5,3.5)
 {\small \underline{Clausal form}:
\footnotesize
\begin{tabular}{l}
\drgvar{b1} \texttt{IMP} \drgvar{b2} \drgvar{b3}\\
\drgvar{b2} \texttt{REF} \drgvar{x1}\\
\drgvar{b2} \texttt{wild} \drgvar{x1}\\
\drgvar{b2} \texttt{dog} \drgvar{x1}\\
\drgvar{b3} \texttt{run} \drgvar{x1}\\
\end{tabular}
};
\end{tikzpicture}
}%scalebox

\caption{Illustration of \todo{our} evaluation protocol using SyGNS.
The goal is to map English sentences to meaning representations.
The generalization test evaluates
novel combinations of operations (modifier, quantifier, negation) in the training set.
We use multiple meaning representations and evaluation methods.}
\label{fig:goodpic}
\end{figure}
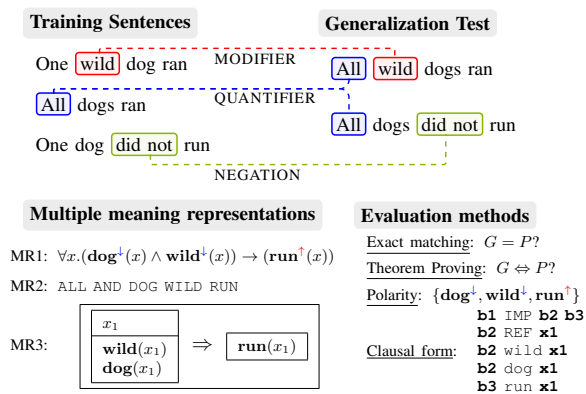

To address these issues,
we propose SyGNS (pronounced as \textit{signs}), a Systematic Generalization testbed based on Natural language Semantics.
The goal is to map English sentences
to various meaning representations,
so it can be taken as a sequence-to-sequence semantic parsing task.

Figure~\ref{fig:goodpic} illustrates our evaluation protocol using SyGNS.
To address the first issue above,
we probe the generalization capability of DNN models on two out-of-distribution tests: \textit{systematicity} (Section~\ref{ssec:unscomp}) and \textit{productivity} (Section~\ref{ssec:unsdep}), 
two concepts treated as hallmarks of human cognitive capacities in cognitive sciences~\cite{Fodor1988-FODCAC,calvo2014architecture}.
We use a train-test split controlled by each target concept and train models with a minimally sized training set (Basic set) involving primitive patterns composed of semantic phenomena such as quantifiers, modifiers, and negation.
If a model learns different properties of each semantic phenomenon from the Basic set,
it should be able to parse a sentence with novel combination patterns.
Otherwise, a model has to memorize an exponential number of combinations of linguistic expressions.

To address the second issue, we use multiple forms of meaning representations developed in formal semantics~\cite{Montague73,Heim-Kratzer98,jacobson2014compositional} and their respective evaluation methods.
\todo{We use three \textit{scoped} meaning representation forms, each of which preserves the same semantic information (Section~\ref{ssec:sr}).
In formal semantics,
it is generally assumed that 
scoped meaning representations are standard forms
for handling diverse semantic phenomena such as quantification and negation.
Scoped meaning representations also enable us
to evaluate the compositional generalization ability of the models 
to capture semantic phenomena in a more fine-grained way.
By decomposing an output meaning representation into constituents (e.g., words) in accordance with its structure, we can compute the matching ratio between the output representation and the gold standard representation.
Evaluating the models on multiple meaning representation forms also allows us to explore whether the performance depends on the complexity of the representation forms.
}

This paper provides three main contributions.
First, we \todo{develop} the SyGNS \todo{testbed} to test model ability to systematically transform sentences involving linguistic phenomena into multiple forms of scoped meaning representations.
\todo{The data and code for SyGNS are publicly available at \url{https://github.com/verypluming/SyGNS}.}
Second, we use SyGNS to analyze the systematic generalization capacity of two standard DNN models: Gated Recurrent Unit (GRU) and Transformer.
Experiments show that these models can generalize to unseen combinations of quantifiers, negations, and modifiers \todo{to some extent.
However, the generalization ability is limited to the combinations whose forms are similar to those of the training instances.}
\todo{In addition, the models} struggle with parsing sentences involving nested clauses.
We also show that the extent of generalization depends on the choice of primitive patterns and representation forms.

\section{Related Work}
\label{sec:related}
The question of whether neural networks
obtain the systematic generalization capacity
has long been discussed~\cite{Fodor1988-FODCAC,962dc7dfb35547148019f194381d2cc6,baroni2020linguistic}.
Recently, empirical studies using NLI tasks have revisited this question, showing that current models learn undesired biases~\cite{glockner-etal-2018-breaking,poliak-etal-2018-hypothesis,tsuchiya-2018-performance,geva-etal-2019-modeling,liu-etal-2019-inoculation} and heuristics~\cite{mccoy-etal-2019-right}, and fail to 
\todo{consistently learn various} inference types~\cite{rozen-etal-2019-diversify,DBLP:conf/aaai/NieWB19,yanaka-etal-2019-neural,Richardson2019,joshi-etal-2020-taxinli}.
In particular, previous works~\cite{goodwin-etal-2020-probing,yanaka-etal-2020-neural,geiger-etal-2020-neural,yanaka-etal-2021-exploring} have examined whether models learn \pending{the systematicity of NLI on monotonicity and veridicality}.
While this line of work has shown certain limitations of model generalization capacity, 
it is often difficult to figure out why the NLI model fails and how to improve it,
partly because NLI tasks depend on multiple factors, including semantic interpretation of target phenomena and acquisition of background knowledge.
By focusing on semantic parsing rather than NLI,
one can probe \pending{to what extent models systematically interpret the meaning of sentences according to their structures and the meanings of their constituents.}

Meanwhile, datasets for analysing the compositional generalization ability of DNN models in semantic parsing have been proposed, including SCAN~\cite{Lake2017GeneralizationWS,baroni2020linguistic}, CLUTRR~\cite{sinha-etal-2019-clutrr}, and CFQ~\cite{keysers2020measuring}.
For example, the SCAN task is to investigate whether models trained with a set of primitive instructions (e.g., \textit{jump} $\rightarrow$ \texttt{JUMP}) and modifiers (e.g., \textit{walk twice} $\rightarrow$ \texttt{WALK WALK}) generalize to new combinations of primitives (e.g., \textit{jump twice} $\rightarrow$ \texttt{JUMP JUMP}).
However, these datasets deal with artificial languages, where the variation of linguistic expressions is limited, so it is not clear to what extent the models
systematically interpret various semantic phenomena in natural language, such as quantification and negation.

Regarding the generalization capacity of DNN models in natural language, previous studies have focused on syntactic and morphological generalization capacities such as subject-verb agreement tasks~\citep[][i.a.]{linzen-etal-2016-assessing,gulordava-etal-2018-colorless,marvin-linzen-2018-targeted}.
Perhaps closest to our work is the COGS task~\cite{kim-linzen-2020-cogs} for probing the generalization capacity of semantic parsing in a synthetic natural language fragment.
For instance, the task is to see whether models trained to parse sentences where some lexical items only appear in subject position (e.g., \textit{\underline{John} ate the meat}) can generalize to structurally related sentences where these items appear in object position (e.g., \textit{The kid liked \underline{John}}).
In contrast to this work, our focus is more on semantic parsing of sentences with logical and semantic phenomena that require scoped meaning representations.
Our study also improves previous work on the compositional generalization capacity in semantic parsing in that we compare three types of meaning representations and evaluate them at multiple levels, including logical entailment, polarity assignment, and partial clause matching (Section \ref{ssec:sr}).

\section{Overview of SyGNS}
\label{sec:method}
We use two evaluation concepts to assess the systematic capability of models: systematicity (Section~\ref{ssec:unscomp}) and productivity (Section~\ref{ssec:unsdep}).
In evaluating these two concepts, we use synthesized pairs of sentences and their meaning representations to control a train-test split (Section~\ref{ssec:gen}).
The main idea is to analyze models trained with a minimum size of a training set (Basic set) involving primitive patterns composed of various semantic phenomena;
if a model systematically learns primitive \todo{combination} patterns in the Basic set,
it should parse a new sentence with different combination patterns.
We target three types of scoped meaning representations and use their respective evaluation methods, according to the function and structure of each representation form (Section~\ref{ssec:sr}).

\subsection{Systematicity}
\label{ssec:unscomp}
Table~\ref{tab:sys} illustrates how we test systematicity,
i.e., the capacity to interpret novel combinations of
primitive semantic phenomena.
We generate Basic set~1 by combining various quantifiers with
sentences without modifiers.
We also generate Basic set~2 by setting an arbitrary quantifier (e.g., \textit{one}) to a primitive quantifier and combining it with various types of modifiers.
We then evaluate whether models trained with Basic sets~1 and 2 can parse sentences involving unseen combinations of quantifiers and modifiers.
We also test the combination of quantifiers and negation in the same manner; the detail is given in Appendix~D.

To provide a controlled setup, we use three quantifier types: existential quantifiers (\Repl{Exi}),
numerals (\Repl{Num}),
and universal quantifiers (\Repl{Uni}).
Each type has two patterns:
\textit{one} and \textit{a} for \Repl{Exi},
\textit{two} and \textit{three} for \Repl{Num},
and \textit{all} and \textit{every} for \Repl{Uni}.
We consider three settings where 
the primitive quantifier is set to
the type \Repl{Exi}, \Repl{Num}, or \Repl{Uni}.

For modifiers, we distinguish three types --- adjectives (\Repl{Adj}), adverbs (\Repl{Adv}), and logical connectives (\Repl{Con}; conjunction \textit{and}, disjunction \textit{or}) --- and ten patterns for each.
Note that each modifier type differs in its position;
an adjective appears inside a noun phrase (e.g., one \textit{small} tiger), while an adverb (e.g., \textit{quickly}) and a coordinated phrase with a logical connective (e.g., \textit{or came}) appears at the end of a sentence.
Although Table~\ref{tab:sys} only shows the pattern generated by the primitive quantifier \textit{one} and the noun \textit{tiger}, the noun can be replaced with ten other nouns (e.g., \textit{dog}, \textit{cat}, etc.) in each setting.
See Appendix A for more details on the fragment of English considered here.

\begin{table}
  \centering
  \scalebox{0.91}{
  \begin{tabular}{lll} \hline
          \multicolumn{2}{l}{Pattern}&Sentence\\ \hline
            \multicolumn{3}{c}{Train} \\ 
            \multicolumn{2}{l}{Primitive quantifier}&\textbf{One} tiger ran\\
             Basic 1&\Repl{Exi}&\textbf{A} tiger ran \\ 
             &\Repl{Num}&\textbf{Two} tigers ran\\ 
             &\Repl{Uni}&\textbf{Every} tiger ran\\
             Basic 2&\Repl{Adj}& \textbf{One} \textit{small} tiger ran \\ 
             &\Repl{Adv}&\textbf{One} tiger ran \textit{quickly}\\
             &\Repl{Con}&\textbf{One} tiger ran \textit{or came}\\ \hline \hline
             \multicolumn{3}{c}{Test} \\ 
             \multicolumn{2}{l}{\Repl{Exi}$+$\Repl{Adj}}&\textbf{A} \textit{small} tiger ran\\
             \multicolumn{2}{l}{\Repl{Num}$+$\Repl{Adv}}&\textbf{Two} tigers ran \textit{quickly}\\
             \multicolumn{2}{l}{\Repl{Uni}$+$\Repl{Con}}&\textbf{Every} tiger ran \textit{or came} \\ \hline
    \end{tabular}
    }
    \caption{Training and test instances for systematicity.}
  \label{tab:sys}
\end{table}

\begin{table}
  \centering
    \scalebox{0.91}{
  \begin{tabular}{llp{11.5em}} \hline
          \multicolumn{2}{l}{Pattern}&Sentence\\ \hline
            \multicolumn{3}{c}{Train (Basic 1: depth 0, Basic 2: depth 1)} \\ 
             Basic 1&\Repl{Non}& Two dogs loved Ann \\ 
             Basic 2&\Repl{Per}&Bob liked a bear [that chased all polite cats]\\ 
             &\Repl{Cen}&Two dogs [that all cats kicked] loved Ann \\ \hline \hline
             \multicolumn{3}{c}{Test (examples: depth 2)} \\ 
             \multicolumn{2}{l}{\Repl{Per}$+$\Repl{Per}}&Bob liked a bear\\
             &&[that chased all polite cats\\
             &&[that loves Ann]]\\
            \multicolumn{2}{l}{\Repl{Per}$+$\Repl{Cen}}&Two dogs [that a bear\\
             &&[that chased all polite cats]\\
             &&kicked] loved Ann\\ \hline
    \end{tabular}
    }
    \caption{Training and test instances for productivity.}
  \label{tab:pro}
\end{table}

\subsection{Productivity}
\label{ssec:unsdep}
Productivity refers to the capacity to interpret
an indefinite number of sentences with recursive operations. 
To test productivity, we use embedded relative clauses,
which interact with quantifiers to generate logically complex sentences.
Table~\ref{tab:pro} shows examples.
We provide two Basic sets;
Basic set~1 consists of sentences without embedded clauses (\Repl{Non}) and
Basic set~2 consists of sentences with a single embedded clause, which we call sentences with depth one.
We then test whether models trained with Basic sets~1 and 2 can parse a sentence involving deeper embedded clauses,
i.e., sentences whose depth is two or more.
As Table~\ref{tab:pro} shows, we consider both peripheral-embedding (\Repl{Per}) and center-embedding (\Repl{Cen}) clauses.

\subsection{Meaning representation and evaluation}
\label{ssec:sr}

\paragraph{Overview}
To evaluate generalization capacity in semantic parsing at multiple levels, we use three types of scoped meaning representations: (i) First-Order Logic (FOL) formulas, (ii) Discourse Representation Structures~\cite[DRSs;][]{Kamp1993-KAMFDT}, and (iii) Variable-Free (VF) formulas~\cite{baader2003description,pratt-hartmann_moss_2009}.
DRSs can be converted to \textit{clausal forms}~\cite{van-noord-etal-2018-evaluating} for evaluation.
For instance, \todo{the} sentence (\ref{ex:sent}) is mapped to the FOL formula in (\ref{ex:fol}),
the DRS in (\ref{ex:drs}a), its clausal form in (\ref{ex:drs}b),
and the VF formula in (\ref{ex:vf}).
\begin{exe}
\ex \label{ex:sent} One white dog did not run.
\ex \label{ex:fol}
$\exists x_1. (\LF{white}(x_1) \wedge \LF{dog}(x_1) \wedge \neg \LF{run}(x_1))$
\ex \label{ex:drs}
a.
\scalebox{0.8}{
\begin{minipage}{8em}
\drs{$x_1$}{$\LF{white}(x_1)$ \\
$\LF{dog}(x_1)$ \\
$\neg$ \drs{}{$\LF{run}(x_1)$}}
\end{minipage}
}%scalebox
\hspace{1em}
b.
\scalebox{0.8}{
\begin{minipage}{5em}
    \begin{tabular}{l}
     \drgvar{b1} \texttt{REF} \drgvar{x1}\\
     \drgvar{b1} \texttt{white} \drgvar{x1}\\
     \drgvar{b1} \texttt{dog} \drgvar{x1}\\
     \drgvar{b1} \texttt{NOT} \drgvar{b2}\\
     \drgvar{b2} \texttt{run} \drgvar{x1}\\
     \end{tabular}
\end{minipage}
}%scalebox
\ex \label{ex:vf}
\texttt{EXIST} \texttt{AND} \texttt{WHITE} \texttt{DOG} \texttt{NOT} \texttt{RUN}
\end{exe}

\noindent
Using these multiple forms enables us to analyze whether the difficulty in semantic generalization depends on the format of meaning representations.

Previous studies for probing generalization capacity in semantic parsing~\cite[e.g.,][]{Lake2017GeneralizationWS,sinha-etal-2019-clutrr,keysers2020measuring,kim-linzen-2020-cogs}
use a fixed type of meaning representation, with its evaluation method limited to the exact-match percentage, where an output is considered correct only if it exactly matches the gold standard. However, this does not properly assess whether models capture the structure and function of meaning representation.
First, exact matching does not directly take into account whether two meanings are logically equivalent~\cite{BlackburnBos05}:
for instance, schematically two formulas $A \wedge B$ and $B \wedge A$ are different in form but have the same meaning.
Relatedly, scoped meaning representations for natural languages can be made complex by including parentheses and variable renaming mechanism (the so-called $\alpha$-conversion in $\lambda$-calculus). For instance, we want to identify two formulas which only differ in variable naming,
e.g., $\exists x_1.F(x_1)$ and $\exists x_2.F(x_2)$.
It is desirable to compare exact matching with
alternative evaluation methods,
and to consider alternative meaning representations 
that avoid these problems.
Having this background in mind, below we will describe each type of meaning representation in detail.

\paragraph{FOL formula}
FOL formulas are standard forms in formal and computational semantics~\cite{BlackburnBos05,Jurafsky:2009:SLP:1214993}, where content words such as nouns and verbs
are represented as predicates, and function words such as quantifiers, negation, and connectives are represented as logical operators with scope relations (cf. the example in (\ref{ex:fol})).
To address the issue on evaluation, we consider
two ways of evaluating FOL formulas in addition to exact matching: (i) automated theorem proving (ATP) and (ii) monotonicity-based polarity assignment.

First, FOL formulas can be evaluated by checking the logical entailment relationships that directly consider whether two formulas are logically equivalent.
Thus we evaluate predicted FOL formulas by using ATP.
We check whether a gold formula $G$ entails prediction $P$ and vice versa, using an off-the-shelf FOL theorem prover\footnote{We use a state-of-the-art FOL prover Vampire available at \url{https://github.com/vprover/vampire}}.
To see the logical relationship between $G$ and $P$,
we measure the accuracy for unidirectional and bidirectional entailment: \righte, \lefte, and \bie.

Second, the polarity of each content word appearing in a sentence can be extracted from the FOL formula
using its monotonicity property~\cite{van1986essays,MacCartney:2007:NLT:1654536.1654575}. This enables us to analyze whether models can correctly capture
entailment relations triggered by quantifier and negation scopes.
Table~\ref{tab:evalmono} shows some examples of monotonicity-based polarity assignments.
For example, existential quantifiers such as \textit{one} 
are upward monotone (shown as \textcolor{red}{$\uparrow$})
with respect to the subject NP and the VP,
because they can be substituted with their hypernyms (e.g.,
\textit{One dog ran} $\Rightarrow$ \textit{One animal moved}).
These polarities can be extracted from the FOL formula because $\exists$ and $\wedge$ are upward monotone operators in FOL.
Universal quantifiers such as \textit{all} are downward monotone (shown as \textcolor{blue}{$\downarrow$}) with respect to the subject NP and upward monotone with respect to the VP. Expressions in downward monotone position can be substituted with their hyponymous expressions (e.g., \textit{All dogs ran} $\Rightarrow$ \textit{All white dogs ran}).
The polarity can be reversed by embedding another downward entailing context like negation, so the polarity of \textit{run} in the third case in Table~\ref{tab:evalmono}
is flipped to downward monotone.\footnote{We follow the surface order of NPs and take it that the subject NP always take scope over the VP.}
For evaluation based on monotonicity, we extract a polarity
for each content word in a gold formula and a prediction and calculate the F-score for each monotonicity direction (upward and downward).

\begin{table}
    \centering
    \scalebox{0.95}{
    \begin{tabular}{ll}
         One \pospol{dog} \pospol{ran}:&{\small $\exists x. (\pospol{\LF{dog}}(x) \wedge \pospol{\LF{run}}(x))$} \\
         All \negpol{dogs} \pospol{ran}:& {\small $\forall x. (\negpol{\LF{dog}}(x)\to\pospol{\LF{run}}(x))$} \\
         All \negpol{dogs} did not \negpol{run}:&{\small $\forall x. (\negpol{\LF{dog}}(x)\to\neg \negpol{\LF{run}}(x))$}\\
    \end{tabular}
    }
    \caption{Examples of monotonicity-based polarity assignments for FOL formulas.}
    \label{tab:evalmono}
\end{table}

\paragraph{DRS}
A DRS is a form of scoped meaning representations proposed in Discourse Representation Theory, a well-studied formalism in formal semantics~\cite{Kamp1993-KAMFDT,Asher1993ReferenceTA,025f4c3cc8834716b7ad62f569ed5a72}.
By translating a box notation as in (\ref{ex:drs}a)
to the clausal form as in  (\ref{ex:drs}b), one can evaluate DRSs by \textsc{Counter}\footnote{\url{https://github.com/RikVN/DRS\_parsing}}, which is a standard tool for evaluating neural DRS parsers~\cite{liu-etal-2018-discourse,van-noord-etal-2018-exploring}.
\textsc{Counter} searches for the best variable mapping between predicted DRS clauses and gold DRS clauses and calculates an F-score over matching clause, which is similar to \textsc{Smatch}~\cite{cai-knight-2013-smatch}, an evaluation metric designed for Abstract Meaning Representation (AMR;~\modelcite{banarescu-etal-2013-abstract}).
\textsc{Counter} alleviates the process of variable renaming and correctly evaluates the cases where the order of clauses is different from that of gold answers.

\paragraph{VF formula}
FOL formulas in our fragment have logically equivalent forms in a variable-free format,
which does not contain parentheses nor variables as in the example (\ref{ex:vf}).
Our format is similar to a variable-free form in Description Logic~\cite{baader2003description} and Natural Logic~\cite{pratt-hartmann_moss_2009}.
VF formulas alleviate the problem of parentheses and variable renaming, while preserving semantic information~\cite[cf.\,][]{NIPS2017_18d10dc6}.
Due to the equivalence with FOL formulas, it is possible to extract polarities from VF formulas.
See Appendix A for more examples of VF formulas.

\begin{table*}
    \centering
    \scalebox{0.9}{
    \begin{tabular}{l|cccc|cccc}\hline
    \multirow{2}{*}{Test}&\multicolumn{4}{c|}{GRU}&\multicolumn{4}{c}{Transformer}\\
    &FOL&DRS&DRS (cnt)&VF&FOL&DRS&DRS (cnt)&VF\\ \hline
    \multicolumn{9}{c}{primitive quantifier: existential quantifier \textit{one}}\\ \hline
    \Repl{Exi}&96.1&99.5&99.9&99.7&99.9&99.8&99.9&100.0\\
    \Repl{Num}&7.6&12.7&86.0&37.0&18.1&96.9&99.7&20.7\\ 
    \Repl{Uni}&3.1&4.4&56.8&39.5&8.3&2.2&74.2&17.7\\ \hline
    Valid&98.2&99.7&100.0&99.6&100.0&100.0&100.0&100.0 \\ \hline
    \multicolumn{9}{c}{primitive quantifier: numeral \textit{two}}\\ \hline
    \Repl{Exi}&11.6&42.1&91.4&45.3&34.0&84.5&98.3&10.5\\
    \Repl{Num}&59.5&83.6&98.7&42.8&99.9&97.4&99.8&80.9\\ 
    \Repl{Uni}&2.5&1.8&68.6&39.2&0.0&0.1&72.3&90.9\\ \hline
    Valid&84.3&99.7&100.0&98.9&100.0&100.0&100.0&100.0\\ \hline
    \multicolumn{9}{c}{primitive quantifier: universal quantifier \textit{every}}\\ \hline
    \Repl{Exi}&1.6&0.3&43.8&61.3&2.1&0.2&70.8&20.8\\
    \Repl{Num}&1.4&0.3&75.9&69.3&0.1&0.1&76.8&99.7\\ 
    \Repl{Uni}&33.8&96.5&99.4&100.0&100.0&100.0&100.0&99.9\\ \hline
    Valid&93.4&100.0&100.0&99.3&100.0&100.0&100.0&99.9\\ \hline
    \multicolumn{9}{c}{primitive quantifiers: \textit{one}, \textit{two}, \textit{every}}\\ \hline
    \Repl{Exi}&99.7&99.0&100.0&100.0&100.0&100.0&100.0&100.0\\
    \Repl{Num}&91.2&96.4&99.2&99.3&100.0&100.0&100.0&100.0\\ 
    \Repl{Uni}&95.7&97.6&99.4&100.0&99.9&100.0&100.0&100.0\\ \hline
    Valid&98.4&100.0&100.0&99.3&100.0&100.0&100.0&100.0\\ \hline
    \end{tabular}
    }
    \caption{Accuracy by quantifier type. ``DRS (cnt)'' columns show the accuracy of predicted DRSs by \textsc{Counter}, and ``Valid'' rows show the validation accuracy. Each accuracy is measured by exact matching, except for ``DRS (cnt)'' columns.}
    \label{tab:quant50Koneall}
\end{table*}

\subsection{Data generation}
\label{ssec:gen}
To provide synthesized data, we generate sentences using a context-free grammar (CFG) 
associated with semantic composition rules in the standard $\lambda$-calculus (see Appendix~A for details).
Each sentence is mapped to an FOL formula and VF formula
by using the semantic composition rules specified in the CFG.
DRSs are converted from the generated FOL formulas using the standard mapping~\cite{Kamp1993-KAMFDT}.
To generate controlled fragments for each train-test split,
the CFG rules automatically annotate the types of semantic phenomena
involved in sentences generated.
We annotate seven types: the positions of quantifiers (subject or object), negation, adjectives, adverbs, conjunction, disjunction, and embedded clause types (peripheral or center embedding).

\todo{To test} systematicity, we generate \todo{sentences using the CFG}, randomly select 50,000 examples, and then split them into 12,000 training examples and 38,000 test examples according to a primitive quantifier.
\todo{To test} productivity, we apply up to four recursive rules and \todo{randomly select} 20,000 examples for each depth.

\section{Experiments and Analysis}
\label{sec:experiment}
Using SyGNS, we test the performance of Gated Recurrent Unit (GRU;~\modelcite{cho-etal-2014-learning}) and Transformer~\cite{NIPS2017_3f5ee243} in an encoder-decoder setup.
These are widely used models to perform well on hierarchical generalization tasks~\cite{DBLP:conf/cogsci/McCoyFL18,russin-etal-2020-compositional}.

\subsection{Experimental setup}
\label{ssec:setup}
\todo{In all experiments, 
we trained each model for 25 epochs with early stopping (patience = 3).}
We performed five runs 
and reported their average accuracies.
The input sentence is represented as a sequence of words, using spaces as a separator.
The maximum input and output sequence length is set to the length of a sequence with maximum depths of embedded clauses.
We set the dropout probability to 0.1 on the output and used a batch size of 128 and an embedding size of 256.
\todo{Since incorporating pre-training would make it hard to distinguish whether the models' ability to perform semantic parsing comes from training data or from pre-training data, we did not use any pre-training.}

For the GRU, we used a single-layer encoder-decoder with global attention and a dot-product score function.
Since a previous work~\cite{kim-linzen-2020-cogs} reported that unidirectional models are more robust regarding sentence structures than bi-directional models, we selected a unidirectional GRU encoder.
For the Transformer, we used a three-layer encoder-decoder, a model size of 512, and a hidden size of 256.
\todo{The number of model parameters was 10M, respectively.}
\todo{See Appendix~B for additional training details.}

\subsection{Results on systematicity}
\label{ssec:excomp}
\paragraph{Generalization on quantifiers}
Table~\ref{tab:quant50Koneall} shows the accuracy by quantifier type.
When the existential quantifier \textit{one} was the primitive quantifier, the accuracy on the problems involving existential quantifiers, which have the same type as the primitive quantifier, was nearly perfect.
Similarly, when the universal quantifier \textit{every} was the primitive quantifier, the accuracy on the problems involving universal quantifiers was much better than that on the problems involving other quantifier types.
\todo{These results indicate that models can easily generalize to problems involving quantifiers of the same type as the primitive quantifier, while the models struggle with generalizing to the others.}
\todo{We also experimented with larger models and observed the same trend (see Appendix~C).}
The extent of generalization varies according to the primitive quantifier type and meaning representation forms.
For example, when the primitive quantifier is the numeral expression \textit{two}, models generalize to problems of VF formulas involving universal quantifiers.
This can be explained by the fact that
VF formulas involving universal quantifiers like (\ref{ex:allf}) have a similar form to those involving numerals as in (\ref{ex:twof}), whereas FOL formulas involving universal quantifiers have a different form from those involving numerals as in (\ref{ex:allfol}) and (\ref{ex:twofol}).

\begin{exe}
\ex 
\begin{xlist}
\ex \label{ex:alls} All small cats chased Bob
\ex \label{ex:allf} {\small\texttt{ALL AND CAT SMALL EXIST BOB CHASE}}
\ex \label{ex:allfol} {\small $\forall x_1.(\LF{cat}(x_1) \wedge \LF{small}(x_1)$}\\
\quad {\small $\to \exists x_2.(\LF{bob}(x_2) \wedge \LF{chase}(x_1,x_2)))$}
\end{xlist}
\ex 
\begin{xlist}
\ex \label{ex:twos} Two small cats chased Bob
\ex \label{ex:twof} {\small\texttt{TWO AND CAT SMALL EXIST BOB CHASE}}
\ex \label{ex:twofol} {\small $\exists x_1.(\LF{two}(x_1) \wedge \LF{cat}(x_1) \wedge \LF{small}(x_1)$}\\
\quad {\small $\wedge \ \exists x_2.(\LF{bob}(x_2) \wedge \LF{chase}(x_1,x_2)))$}
\end{xlist}

\end{exe}

\noindent \todo{We also check the performance when three quantifiers \textit{one}, \textit{two}, and \textit{every} are set as primitive quantifiers.
This setting is easier than that for the systematicity in Table~\ref{tab:sys}, since the models are exposed to combination patterns of all the quantifier types and all the modifier types.
In this setting, the models achieved almost perfect performance on the test set involving non-primitive quantifiers (\textit{a}, \textit{three}, \textit{all}).
}

\begin{table*}
    \centering
    \scalebox{0.88}{
    \begin{tabular}{l|cccc|cccc}\hline
    \multirow{2}{*}{Test}&\multicolumn{4}{c|}{GRU}&\multicolumn{4}{c}{Transformer}\\
    &FOL&DRS&DRS (cnt)&VF&FOL&DRS&DRS (cnt)&VF\\ \hline
    \Repl{Adj}&18.9&20.1&78.1&42.3&26.8&59.1&91.3&27.6\\
    \Repl{Adj}$+$\Repl{Neg}&18.8&20.2&80.5&39.7&23.1&59.5&93.6&27.5\\ \hline
    \Repl{Adv}&20.1&47.7&87.5&58.4&36.2&67.3&97.6&50.7\\ 
    \Repl{Adv}$+$\Repl{Neg}&26.9&62.7&92.7&67.2&50.7&69.4&97.3&62.1\\ \hline
    \Repl{Con}&28.9&58.3&84.7&72.9&54.3&66.8&88.3&65.9\\ 
    \Repl{Con}$+$\Repl{Neg}&33.6&62.8&86.6&74.9&60.1&65.1&89.9&69.1\\ \hline
    Valid&98.2&99.7&100.0&99.6&100.0&100.0&100.0&100.0 \\ \hline
    \end{tabular}
    }
    \caption{Accuracy by modifier type (primitive quantifier: existential quantifier \textit{one}). $+$\Repl{Neg} indicates problems involving negation. Each accuracy is measured by exact matching, except for ``DRS (cnt)'' columns.}
    \label{tab:pheno50K}
\end{table*}

\begin{figure*}
\begin{minipage}{0.25\hsize}
\scalebox{0.33}{\includegraphics[bb=0.000000 0.000000 357.630995 303.866335]{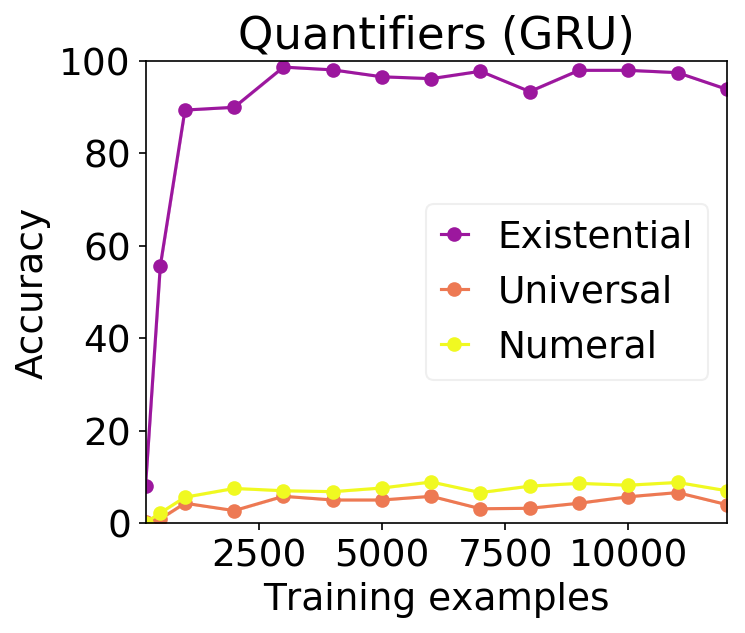}}
\end{minipage}
\begin{minipage}{0.24\hsize}
\scalebox{0.33}{\includegraphics[bb=0.000000 0.000000 357.630995 303.866335]{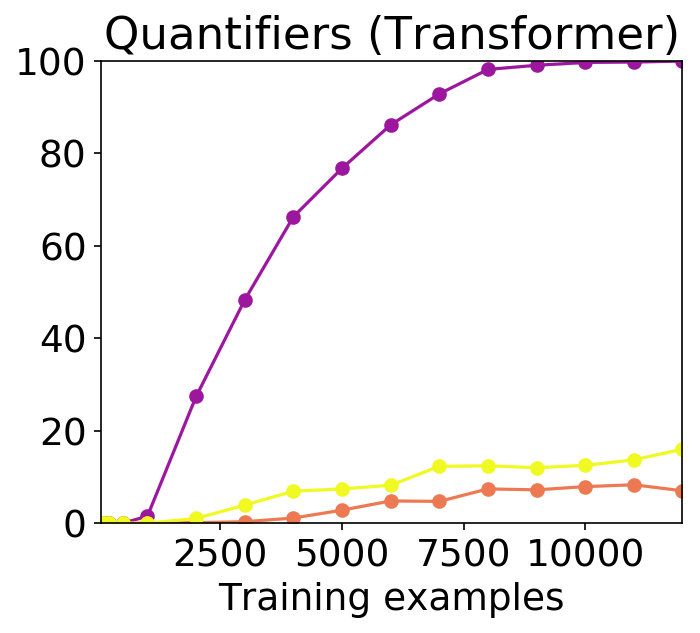}}
\end{minipage}
\begin{minipage}{0.24\hsize}
\scalebox{0.33}{\includegraphics[bb=0.000000 0.000000 357.630995 303.866335]{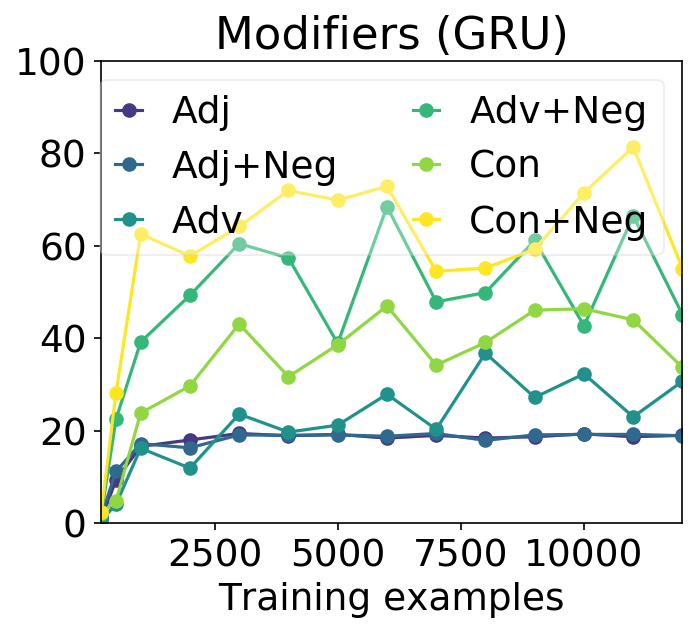}}
\end{minipage}
\begin{minipage}{0.24\hsize}
\scalebox{0.33}{\includegraphics[bb=0.000000 0.000000 357.630995 303.866335]{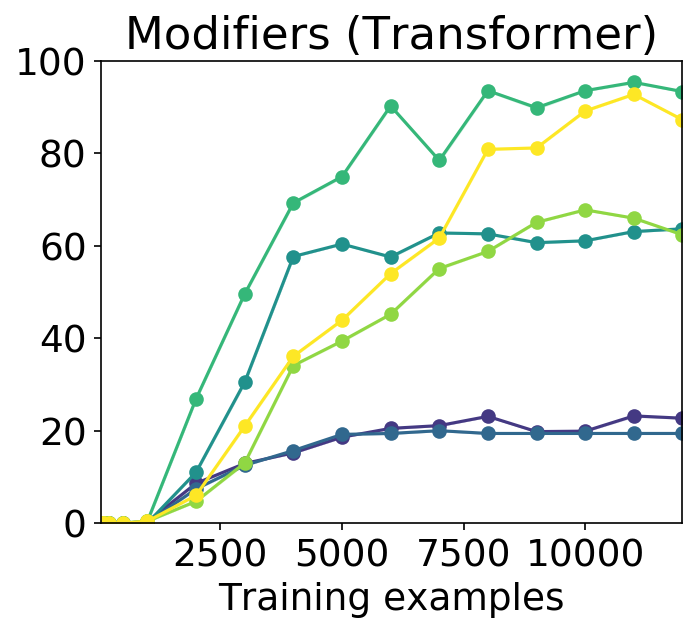}}
\end{minipage}
\caption{Learning curves on FOL formula generalization tasks (primitive quantifier: \textit{one}).}
\label{fig:lc}
\end{figure*}

\paragraph{Generalization on modifiers}
Table~\ref{tab:pheno50K} shows the accuracy by modifier type where \textit{one} is set to the primitive quantifier (see Appendix~C for the results where other quantifier types are set to the primitive quantifier).
No matter which quantifier is set as the primitive quantifier, the accuracy for problems involving logical connectives or adverbs is better than those involving adjectives.
As in (\ref{ex:adjs}), an adjective is placed between a quantifier and a noun, so the position of the noun \textit{dog} with respect to the quantifier \textit{every} in the test set changes from the example in the training (Basic) set in (\ref{ex:s}).
In contrast, adverbs and logical connectives are placed at the end of a sentence, so the position of the noun does not change from the training set, as in (\ref{ex:cons}).
This suggests that models can more easily generalize in problems involving unseen combinations of quantifiers and modifiers where the position of the noun is the same between the training and test sets.

\begin{exe}
\ex \label{ex:s} Every \underline{dog} ran
\hfill{Train (Basic set)}
\ex \label{ex:adjs} Every \underline{large dog} ran
\hfill{Test (\Repl{Adj})}
\ex \label{ex:cons} Every \underline{dog} ran and cried
\hfill{Test (\Repl{Con})}
\end{exe}

\noindent Table~\ref{tab:pheno50K} also shows that the accuracy is nearly the same regardless of the existence of negation. 
Basic set contains examples involving negation, and this indicates that the existence of complex phenomena like negation does not affect generalization performance of models on modifiers so long as such phenomena are included in the training set.

\begin{table}
    \centering
    \scalebox{0.77}{
    \begin{tabular}{l|ccc|ccc}\hline
    \multirow{3}{*}{Test}&\multicolumn{3}{c|}{GRU}&\multicolumn{3}{c}{Transformer}\\
    &{\small\righte}&{\small\lefte}&{\small\bie}&{\small\righte}&{\small\lefte}&{\small\bie}\\ \hline
    \Repl{Exi}&99.8&100.0&99.8&100.0&100.0&100.0 \\   
    \Repl{Num}&77.1&19.0&10.4&
    91.3&21.1&12.4 \\ 
    \Repl{Uni}&7.1&18.7&2.7&
    21.1&83.4&12.3 \\
    \hline
    \end{tabular}
    }
    \caption{ATP-based evaluation results on FOL formulas (primitive quantifier: \textit{one}).}
    \label{tab:allquantrte}
\end{table}

\begin{table}
\centering
    \scalebox{0.80}{
    \begin{tabular}{ll}\hline
    Input & Every wild cat escaped and ran\\
    Gold & {\small $\forall x.((\negpol{\LF{cat}}(x)\wedge \negpol{\LF{wild}}(x))\to (\pospol{\LF{escape}}(x)\wedge \pospol{\LF{run}}(x)))$}\\
    GRU & {\small $\forall x.(\negpol{\LF{cat}}(x)\to (\pospol{\LF{escape}}(x)\wedge \pospol{\LF{run}}(x)))$}\\
    Trans & {\small $\forall x.(\negpol{\LF{cat}}(x)\to \pospol{\LF{wild}}(x) \wedge (\pospol{\LF{escape}}(x)\wedge \pospol{\LF{run}}(x)))$}\\ \hline
    \end{tabular}
    }
    \caption{Examples of typical errors.}
    \label{tab:error}
\end{table}

\begin{table}[h!]
    \centering
    \scalebox{0.85}{
    \begin{tabular}{l|cc|cc}\hline
    \multirow{2}{*}{Test}&\multicolumn{2}{c|}{GRU}&\multicolumn{2}{c}{Transformer}\\
    &Up&Down&Up&Down\\ \hline
    \Repl{Exi}&99.9&100.0&100.0&100.0 \\   
    \Repl{Num}&84.8&96.8&88.1&97.5 \\ 
    \Repl{Uni}&90.9&40.7&94.9&73.4 \\ \hline
    \end{tabular}
    }
    \caption{Monotonicity-based evaluation results on FOL (primitive quantifier: \textit{one}). ``Up'' and ``Down'' columns show upward and downward accuracy, respectively.}
    \label{tab:allquantmono}
\end{table}

\begin{table*}[h!]
    \centering
    \scalebox{0.80}{
    \begin{tabular}{l|cccc|cccc}\hline
    \multirow{2}{*}{Test}&\multicolumn{4}{c|}{GRU}&\multicolumn{4}{c}{Transformer}\\
    &FOL&DRS&DRS (cnt)&VF&FOL&DRS&DRS (cnt)&VF\\ \hline
    Dep2&0.36&0.41&55.5&0.32&0.61&0.61&64.6&0.58\\ 
    Dep3&0.04&0.07&45.6&0.04&0.11&0.12&46.6&0.12\\
    Dep4&0.00&0.01&38.0&0.00&0.02&0.02&37.6&0.02\\ \hline
    Valid&100.0&100.0&100.0&100.0&100.0&100.0&100.0&100.0\\
    \hline
    \end{tabular}
    }
    \caption{Accuracy for productivity. ``Dep'' rows show embedding depths, ``DRS (cnt)'' columns show accuracy of predicted DRSs by \textsc{Counter}, and ``Valid'' row shows the validation accuracy. Each accuracy is measured by exact matching, except for ``DRS (cnt)'' columns.}
    \label{tab:depth}
\end{table*}

\begin{table*}[h]
    \centering
    \scalebox{0.80}{
    \begin{tabular}{l|cccc|cccc}\hline
    \multirow{2}{*}{Test}&\multicolumn{4}{c|}{GRU}&\multicolumn{4}{c}{Transformer}\\
    &FOL&DRS&DRS (cnt)&VF&FOL&DRS&DRS (cnt)&VF\\ \hline
    Dep1&22.1&21.9&81.9&48.9&96.6&77.1&96.5&97.5\\
    Dep2&3.52&3.89&59.1&12.3&76.3&54.6&90.5&85.4\\ 
    Dep3&0.15&0.12&43.3&0.31&24.9&4.5&70.4&37.0\\
    Dep4&0.08&0.15&37.7&0.46&4.4&1.6&60.3&5.57\\ \hline
    Valid&94.3&94.9&100.0&96.0&97.6&98.1&100.0&97.8\\
    \hline
    \end{tabular}
    }
    \caption{Evaluation results for systematicity involving embedding quantifiers. ``Dep'' rows show embedding depths.}
    \label{tab:depth2}
\end{table*}

\paragraph{Meaning representation comparison}
Comparing forms of meaning representations, accuracy by exact matching is highest in the order of VF formulas, DRS clausal forms, and FOL formulas. 
This indicates that models can more easily generalize to unseen combinations where the form of meaning representation is simpler;
VF formulas do not contain parentheses nor variables,
DRS clausal forms contain variables but not parentheses, and
FOL formulas contain both parentheses and variables.

\subsection{Model comparison}
Regarding the generalization capacity of models for decoding meaning representations,
the left two figures in Figure~\ref{fig:lc} show learning curves on FOL prediction tasks by quantifier type.
While GRU achieved perfect performance on the same quantifier type as the primitive quantifier, where the number of training data is 2,500, Transformer achieved the same performance
when the number of training data is 8,000.
The right two figures in Figure~\ref{fig:lc} show learning curves by modifier type.
The GRU accuracy is unstable even when the number of training examples is maximal.
In contrast,
the Transformer accuracy is stable when the number of training data exceeds 8,000.
These results indicate that GRU generalizes to unseen combinations of quantifiers and modifiers with a smaller training set than can Transformer, while the Transformer performance is more stable than that of GRU.

\paragraph{ATP-based evaluation}
Table~\ref{tab:allquantrte} shows the ATP-based evaluation results on FOL formulas.
For combinations involving numerals, both GRU and Transformer achieve high accuracies for \righte\ entailments but low accuracies for \lefte\ entailments.
Since both models fail to output the formulas corresponding to modifiers, they fail to prove \lefte\ entailments.
Regarding combinations involving universal quantifiers,
the GRU accuracy for both \righte\ and \lefte\ entailments is low, and the Transformer accuracy for \lefte\ entailments is much higher than that for \righte\ entailments.
As indicated by examples shown in Table~\ref{tab:error}, GRU tends to fail to output the formula for a modifier, e.g., $\LF{wild}(x)$ in this case, while Transformer fails to correctly output the position of the implication ($\to$).
The ATP-based evaluation results reflect
such differences between error trends of models in 
problems involving different forms of quantifiers.

\paragraph{Monotonicity-based evaluation}
Table~\ref{tab:allquantmono} shows 
accuracies for the monotonicity-based polarity assignment evaluation on FOL formulas.
The accuracies were higher than those using exact matching (cf.\,Table~\ref{tab:quant50Koneall}).
Monotonicity-based evaluation captures the polarities assigned to content words even for the problems that exact-matching judges as incorrect because of the differences in form.
Table~\ref{tab:error} shows examples of predicted polarity assignments.
Here both models predicted correct polarities for three content words, \negpol{\LF{cat}}, \pospol{\LF{escape}}, \pospol{\LF{run}}.
Exact-matching cannot take into account such partial matching.
The downward monotone accuracies for problems involving universal quantifiers are low ($40.7$ and $73.4$ in Table \ref{tab:allquantmono}).
In Table \ref{tab:error}, both models
failed to predict the downward monotonicity of \negpol{\LF{wild}}.
The results indicate that both models struggle with capturing the scope of universal quantifiers.
Appendix~C shows the evaluation results on the polarities of VF formulas.

\subsection{Results on productivity}
\label{ssec:exdep}
Table~\ref{tab:depth} shows
very low generalization accuracy for both GRU and Transformer at unseen depths.
Although the evaluation results using \textsc{Counter} on DRS prediction tasks is much higher than those by exact matching,
this is due to the fact that \textsc{Counter} uses partial matching;
both models tended to correctly predict
the clauses in the subject NP
that are positioned at the beginning of the sentence (see Appendix~E for details).

We checked whether models can generalize to unseen combinations involving embedded clauses when the models are exposed to a part of training instances at each depth.
We provide Basic set~1 involving non-embedding patterns like (\ref{ex:s1}), where \textbf{Q} can be replaced with any quantifier.
This Basic set~1 exposes models to all quantifier patterns.
We also expose models to Basic set~2 involving three primitive quantifiers (\textit{one}, \textit{two}, and \textit{every}) at each embedding depth, like (\ref{ex:s2}) and (\ref{ex:s3}).
We provide around 2,000 training instances at each depth.
We then test models on a test set involving the other quantifiers (\textit{a}, \textit{three}, and \textit{all}) at each embedding depth, like (\ref{ex:s4}) and (\ref{ex:s5}).
If models can distinguish quantifier types during training, they can correctly compose meaning representations involving different combinations of multiple quantifiers.
Note that this setting is easier than that for productivity in Table~\ref{tab:pro}, in that models are exposed to some instances at each depth.

\begin{exe}
\ex \label{ex:s1} \textbf{Q} dog(s) liked Bob
\ex \label{ex:s2} \textbf{One} dog liked Bob [that loved \textbf{two} rats]
\ex \label{ex:s3} \textbf{One} dog liked Bob [that loved \textbf{two} rats [that knew \textbf{every} pig]]
\ex \label{ex:s4} \textbf{A} dog liked Bob [that loved \textbf{three} rats]
\ex \label{ex:s5} \textbf{A} dog liked Bob [that loved \textbf{three} rats [that knew \textbf{all} pigs]]
\end{exe}

Table~\ref{tab:depth2} shows that both models partially generalize to the cases where \todo{the} depth is 1 or 2.
However, both models fail to generalize to the cases where \todo{the} depth is 3 or more.
This suggests that even if models are trained with some instances at each depth, the models fail to learn distinctions between different quantifier types and struggle with parsing sentences whose embedding depth is 3 or more.

\section{Conclusion}
\label{sec:conc}
We have introduced an analysis method using SyGNS, a testbed for diagnosing the systematic generalization capability of DNN models in semantic parsing.
\todo{We found that GRU and Transformer generalized to unseen combinations of semantic phenomena whose meaning representations are similar in forms to those in a training set, while the models struggle with generalizing to the others.
}
\todo{In addition,} these models failed to generalize to cases involving nested clauses.
Our analyses using multiple meaning representation and evaluation methods also revealed detailed behaviors of models.
We believe that SyGNS serves as an effective testbed for investigating the ability
to capture compositional meanings in natural language.

\section*{Acknowledgement}
We thank the three anonymous reviewers for their helpful comments and suggestions.
This work was partially supported by JSPS KAKENHI Grant Number JP20K19868.

\bibliographystyle{acl_natbib}
\bibliography{anthology,acl2021}

\clearpage
\appendix

\section{Data generation details}
Table~\ref{tab:lexicon} shows a set of context-free grammar rules and semantic composition rules we use to generate a fragment of English
annotated with meaning representations in the SyGNS dataset.
Each grammar rule is associated with two kinds of semantic composition rules formulated in $\lambda$-calculus.
One is for deriving first-order logic (FOL) formulas, and the other is for deriving variable-free (VF) formulas.
For FOL, semantic composition runs in the standard Montagovian fashion where all NPs (including proper nouns) are treated as generalized quantifiers~\cite{Heim-Kratzer98,jacobson2014compositional}. From FOL formulas, we can extract the \textit{polarity} of each content word using the monotonicity calculus~\cite{vaneijck2005natural}.
Table \ref{tab:formulas} shows some examples of polarized FOL formulas.
The derivation of VF formulas runs in two steps. To begin with,
a sentence is mapped to a variable-free form by semantic composition rules.
For instance, the sentence \textit{a small dog did not swim} is mapped to a variable-free formula 
\texttt{EXIST}(\texttt{AND}(\texttt{SMALL},\texttt{DOG}),\texttt{NOT}(\texttt{SWIM})) by the rules in Table \ref{tab:lexicon}.
Second, since this form is in prefix notation,
all brackets can be eliminated without causing ambiguity.
This produces the resulting VF formula 
\texttt{EXIST} \texttt{AND} \texttt{SMALL} \texttt{DOG} \texttt{NOT} \texttt{SWIM}.
Some other examples are shown in Table \ref{tab:formulas}.
DRSs are converted from FOL formulas in the standard way~\cite{Kamp1993-KAMFDT}.

\section{Training details}
We implemented the GRU model and the Transformer model using PyTorch.
Both models were optimized using Adam~\cite{Adam} at an initial learning rate of 0.0005.
\todo{The hyperparameters (batch size, learning rate, number of epochs, hidden units, and dropout probability) were tuned by random search.}
In all experiments, we trained models on eight NVIDIA DGX-1 Tesla V100 GPUs.
The runtime for training each model was about 1-4 hours, depending on the size of the training set.
The order of training instances was shuffled for each model.
We used 10\% of the training set for a validation set.

\section{Detailed evaluation results}

\paragraph{\todo{Effect of Model Size}}
\todo{The results we report are from a model with 10M parameters.
How does the number of parameters affect the systematic generalization performance of models?
Table~\ref{tab:modelsize} shows the performance of three models of varying size (large: 27M, medium: 10M, small: 4M).
The number of parameters did not have a large impact on the generalization performance; all runs of the models achieved higher than 90\% accuracy on the validation set and the test set involving quantifiers of the same type as the primitive quantifier, while they did not work well on the test set involving the other types of quantifiers. }

\begin{table}
    \centering
    \scalebox{0.8}{
    \begin{tabular}{l|ccc|ccc}\hline
    \multirow{2}{*}{Test}&\multicolumn{3}{c|}{GRU}&\multicolumn{3}{c}{Transformer}\\
    &4M&10M&27M&4M&10M&27M\\ \hline
    \Repl{Exi}&96.8&99.9&97.1&99.9&99.8&99.3\\
    \Repl{Num}&7.1&11.5&10.4&12.3&12.2&12.4\\
    \Repl{Uni}&6.0&4.9&2.9&7.8&5.9&7.9\\ \hline
    Valid&97.2&99.9&97.6&100.0&99.8&97.2\\ \hline
    \end{tabular}
    }
    \caption{\todo{The effect of model size on generalization performance (primitive quantifier: existential quantifier \textit{one}, representation form: FOL).}}
    \label{tab:modelsize}
\end{table}

\paragraph{Modifier type}
Table~\ref{tab:pheno50Kevery} shows all evaluation results by modifier types where \textit{two} or \textit{every} is set to the primitive quantifier.
Regardless of primitive quantifier type, accuracies for problems involving logical connectives or adverbs were better than those for problems involving adjectives.

\paragraph{Monotonicity}
Table~\ref{tab:allquantmonoall} shows all evaluation results of predicted FOL formulas and VF formulas based on monotonicity.
We evaluate the precision, recall, and 
F-score for each monotonicity direction (upward and downward).
Regardless of meaning representation forms, downward monotone accuracy on problems involving universal quantifiers is low.
This indicates that both models struggle with learning the scope of universal quantifiers.

\begin{table*}
    \centering
    \scalebox{0.69}{
    \begin{tabular}{p{0.4cm}llll} \hline
    \multicolumn{3}{l}{\textbf{Grammar rules}} & \textbf{Semantic composition rules: FOL} & \textbf{Semantic composition rules: VF} \\ \hline
    \catS & $\rightarrow$ &\catNP \ \catVP
    & $\sem{\catS} = \sem{\catNP}(\sem{\catVP})$
    & $\sem{\catS} = \sem{\catNP}(\sem{\catVP})$ \\
    \catS & $\rightarrow$ &\catNP \ \textit{did not}\ \catVP
    & $\sem{\catS} = \sem{\catNP}(\lambda x.\neg\sem{\catVP}(x))$
    & $\sem{\catS} = \sem{\catNP}(\texttt{NOT}(\sem{\catVP}))$ \\

    \catNP & $\rightarrow$ &\catPN
    & $\sem{\catNP} = \sem{\catPN}$
    & $\sem{\catNP} = \sem{\catPN}$ \\
    \catNP & $\rightarrow$ &\catQ\ \catN
    & $\sem{\catNP} = \sem{\catQ}(\sem{\catN})$ 
    & $\sem{\catNP} = \sem{\catQ}(\sem{\catN})$ \\
    \catNP & $\rightarrow$ &\catQ\ \catAdj\ \catN
    & $\sem{\catNP} = \sem{\catQ}(\lambda x.(\sem{\catN}(x) \wedge \sem{\catAdj}(x)))$
    & $\sem{\catNP} = \sem{\catQ}(\texttt{AND}(\sem{\catN},\sem{\catAdj}))$ \\
    \catNP & $\rightarrow$ &\catQ \ \catN \ \catSbar
    & $\sem{\catNP} =\sem{\catQ}(\lambda x.(\sem{\catN}(x) \wedge \sem{\catSbar}(x)))$
    & $\sem{\catNP} = \sem{\catQ}(\texttt{AND}(\sem{\catN},\sem{\catSbar}))$ \\ 
    \catVP & $\rightarrow$ &\catIV
    & $\sem{\catVP} = \sem{\catIV}$
    & $\sem{\catVP} = \sem{\catIV}$\\
    \catVP & $\rightarrow$ &\catIV \ \catAdv
    & $\sem{\catVP} = \lambda x.(\sem{\catIV}(x) \wedge \sem{\catAdv}(x))$
    & $\sem{\catVP} = \texttt{AND}(\sem{\catIV},\sem{\catAdv}))$ \\
    \catVP & $\rightarrow$ &\catIV \ or \catIVdash
    & $\sem{\catVP} = \lambda x. (\sem{\catIV}(x) \vee \sem{\catIV'}(x))$
    & $\sem{\catVP} = \texttt{OR}(\sem{\catIV},\sem{\catIV'}))$ \\    
    \catVP & $\rightarrow$ &\catIV \ and \catIVdash
    & $\sem{\catVP} = \lambda x. (\sem{\catIV}(x) \wedge \sem{\catIV'}(x))$
    & $\sem{\catVP} = \texttt{AND}(\sem{\catIV},\sem{\catIV'}))$ \\    
    \catVP & $\rightarrow$ &\catTV \ \catNP
    & $\sem{\catVP} = \lambda x. \sem{\catNP}(\lambda y. \sem{\catTV}(x,y))$
    & $\sem{\catVP} = \sem{\catNP}(\sem{\catTV})$\\

    \catSbar & $\rightarrow$ &\textit{that} \catVP
    & $\sem{\catSbar} = \sem{\catVP}$
    & $\sem{\catSbar} = \sem{\catVP}$\\
    \catSbar & $\rightarrow$ &\textit{that did not} \catVP
    & $\sem{\catSbar} = \lambda x.\neg \sem{\catVP}(x)$
    & $\sem{\catSbar} = \texttt{NOT}(\sem{\catVP})$\\
    \catSbar & $\rightarrow$ &\textit{that} \catNP \ \catTV
    & $\sem{\catSbar} = \lambda y. \sem{\catNP}(\lambda x. \sem{\catTV}(x,y))$
    & $\sem{\catSbar} = \sem{\catNP}(\texttt{INV}(\sem{\catTV}))$ \\
    \catSbar & $\rightarrow$ &\textit{that} \catNP \ \textit{did not} \catTV
    & $\sem{\catSbar} = \lambda y. \sem{\catNP}(\lambda x. \neg \sem{\catTV}(x,y))$
    & $\sem{\catSbar} = \sem{\catNP}(\texttt{NOT}(\texttt{INV}(\sem{\catTV})))$
    \\ \hline

    %\multicolumn{3}{c}{\textbf{Lexicon}} \\ 
    \catQ & $\rightarrow$&\{\textit{every, all, a, one, two, three}\}
    & $\sem{\textit{every}} = \sem{\textit{all}} = \lambda F \lambda G. \forall x (F(x) \to G(x))$
    & $\sem{\textit{every}} = \sem{\textit{all}} = \lambda F \lambda G. \texttt{ALL}(F,G)$ \\
    &&& $\sem{\textit{a}} = \sem{\textit{one}} = \lambda F \lambda G. \exists x. (F(x) \wedge G(x))$
    & $\sem{\textit{a}} = \sem{\textit{one}} = \lambda F \lambda G. \texttt{EXIST}(F,G)$ \\
    &&& $\sem{\textit{two}} = \lambda F \lambda G. \exists x. (\LF{two}(x) \wedge F(x) \wedge G(x))$
    & $\sem{\textit{two}} = \lambda F \lambda G. \texttt{TWO}(F,G)$ \\
    &&& $\sem{\textit{three}} = \lambda F \lambda G. \exists x. (\LF{three}(x) \wedge F(x) \wedge G(x))$
    & $\sem{\textit{three}} = \lambda F \lambda G. \texttt{THREE}(F,G)$ \\
    \catN &$\rightarrow$&\{\textit{dog, rabbit, cat, bear, tiger,...}\}
    & $\sem{\textit{dog}} = \lambda x.\LF{dog}(x)$
    & $\sem{\textit{dog}} = \texttt{DOG}$\\
    \catPN &$\rightarrow$&\{\textit{ann, bob, fred, chris, eliott,...}\}
    & $\sem{\textit{ann}} = \lambda F.F(\LF{ann})$
    & $\sem{\textit{ann}} = \lambda F.\texttt{EXIST}(\texttt{ANN},F)$\\    
    \catIV &$\rightarrow$&\{\textit{ran, walked, swam, danced, dawdled,...}\}
    & $\sem{\textit{ran}} = \lambda x.\LF{run}(x)$
    & $\sem{\textit{ran}} = \texttt{RUN}$\\    
    \catIVdash &$\rightarrow$&\{\textit{laughed, groaned, roared, screamed,...}\}
    & $\sem{\textit{laugh}} = \lambda x.\LF{laugh}(x)$
    & $\sem{\textit{laugh}} = \texttt{LAUGH}$\\
    \catTV &$\rightarrow$&\{\textit{kissed, kicked, cleaned, touched,...}\}
    & $\sem{\textit{kissed}} = \lambda y \lambda x.\LF{kiss}(x,y)$
    & $\sem{\textit{kissed}} = \texttt{KISS}$\\    
    \catAdj &$\rightarrow$&\{\textit{small, large, crazy, polite, wild,...}\}
    & $\sem{\textit{small}} = \lambda x.\LF{small}(x)$
    & $\sem{\textit{small}} = \texttt{SMALL}$ \\
    \catAdv &$\rightarrow$&\{\textit{slowly, quickly, seriously, suddenly,...}\}
    & $\sem{\textit{slowly}} = \lambda x.\LF{slowly}(x)$
    & $\sem{\textit{slowly}} = \texttt{SLOWLY}$ \\
    \hline
    \end{tabular}
    }
    \caption{A set of context-free grammar rules and semantic composition rules used to generate the SyGNS dataset.}
    \label{tab:lexicon}
\end{table*}

\begin{table*}
\centering
\scalebox{0.8}{
\begin{tabular}{lll} \\ \hline
\textbf{Sentence} & \textbf{FOL} & \textbf{VF} \\ \hline
a small dog did not swim
& $\exists x. (\pospol{\LF{small}}(x) \wedge \pospol{\LF{dog}}(x) \wedge \neg \negpol{\LF{swim}}(x))$
& \texttt{EXIST} \texttt{AND} \pospol{\texttt{SMALL}} \pospol{\texttt{DOG}} \texttt{NOT} \negpol{\texttt{SWIM}} \\
all tigers ran or swam
& $\forall x. (\negpol{\LF{tiger}}(x) \to \pospol{\LF{run}}(x) \vee \pospol{\LF{swim}}(x))$
& \texttt{ALL} \negpol{\texttt{TIGER}} \texttt{OR} \pospol{\texttt{RUN}} \pospol{\texttt{SWIM}} \\
ann did not chase two dogs
& $\neg \exists x. (\LF{two}(x) \wedge \negpol{\LF{dog}}(x) \wedge \negpol{\LF{chase}}(\LF{ann}, x))$
& \texttt{EXIST} \texttt{ANN} \texttt{NOT} \texttt{TWO} \negpol{\texttt{DOG}} \negpol{\texttt{CHASE}} \\ \hline
\end{tabular}
}%scalebox
\caption{Example of (polarized) FOL formulas and VF formulas.}
\label{tab:formulas}
\end{table*}

\begin{table*}
    \centering
    \scalebox{0.70}{
    \begin{tabular}{l|cccc|cccc}\hline
    \multirow{2}{*}{Test}&\multicolumn{4}{c|}{GRU}&\multicolumn{4}{c}{Transformer}\\
    &FOL&DRS&DRS (cnt)&VF&FOL&DRS&DRS (cnt)&VF\\ \hline
    \multicolumn{9}{c}{primitive quantifier: numeral \textit{two}}\\ \hline
    \Repl{Adj}&10.7&16.8&74.9&22.8&34.4&58.2&91.6&52.0\\
    \Repl{Adj}$+$\Repl{Neg}&10.1&22.3&79.7&24.3&33.4&58.7&94.4&51.0\\ \hline
    \Repl{Adv}&12.8&29.3&79.8&46.9&40.3&56.1&89.1&60.5\\ 
    \Repl{Adv}$+$\Repl{Neg}&14.1&33.9&83.8&58.6&34.4&56.4&93.3&65.8\\ \hline
    \Repl{Con}&18.3&37.9&80.2&64.8&34.3&52.4&83.4&67.2\\ 
    \Repl{Con}$+$\Repl{Neg}&24.4&40.6&82.6&68.7&31.3&50.8&88.0&68.5\\ \hline
    \multicolumn{9}{c}{primitive quantifier: universal quantifier \textit{every}}\\ \hline
    \Repl{Adj}&7.7&19.5&70.6&58.5&20.7&20.5&89.9&62.2\\
    \Repl{Adj}$+$\Repl{Neg}&6.9&19.2&75.5&56.8&20.3&20.2&92.2&63.6\\ \hline
    \Repl{Adv}&9.2&18.2&82.3&70.1&19.7&19.7&85.1&70.4\\ 
    \Repl{Adv}$+$\Repl{Neg}&14.8&18.1&79.4&76.1&22.7&19.8&89.3&75.5\\ \hline
    \Repl{Con}&14.7&18.0&70.3&79.6&21.5&19.2&68.8&75.8\\ 
    \Repl{Con}$+$\Repl{Neg}&18.3&18.2&80.1&81.2&22.7&19.1&80.5&76.8\\ \hline
    \end{tabular}
    }
    \caption{Accuracy by modifier type where \textit{two} or \textit{every} is the primitive quantifier. ``DRS (cnt)'' columns show accuracies of predicted DRSs by \textsc{Counter}.}
    \label{tab:pheno50Kevery}
\end{table*}

\begin{table*}
    \centering
    \scalebox{0.70}{
    \begin{tabular}{l|ccccccc|ccccccc}\hline
    \multirow{3}{*}{Test}&\multicolumn{7}{c|}{GRU}&\multicolumn{7}{c}{Transformer}\\
    &Exact&\multicolumn{3}{c}{Upward}&\multicolumn{3}{c|}{Downward}&Exact&\multicolumn{3}{c}{Upward}&\multicolumn{3}{c}{Downward}\\
    &Match&Prec&Rec&F&Prec&Rec&F&Match&Prec&Rec&F&Prec&Rec&F\\ \hline
    \multicolumn{15}{c}{FOL formula}\\ \hline
    \Repl{Exi}&96.1&100.0&99.9&99.9&100.0&100.0&100.0&
    99.9&100.0&100.0&100.0&100.0&100.0&100.0 \\   
    \Repl{Num}&7.6&99.2&75.5&84.8&99.6&94.7&96.8&
    18.1&100.0&79.5&88.1&100.0&95.6&97.5 \\ 
    \Repl{Uni}&3.1&92.6&89.9&90.9&42.9&39.4&40.7&
    8.3&97.3&93.4&94.9&79.4&70.0&73.4 \\ \hline
    \multicolumn{15}{c}{VF formula}\\ \hline
    \Repl{Exi}&99.7&100.0&99.9&99.9&99.9&99.9&99.9&
    100.0&100.0&100.0&100.0&100.0&100.0&100.0 \\   
    \Repl{Num}&37.0&70.2&54.4&59.8&68.6&58.5&62.0&
    20.7&99.2&77.0&85.4&99.3&95.4&97.0 \\ 
    \Repl{Uni}&39.5&91.1&80.7&84.4&49.0&35.2&39.7&
    17.7&99.9&97.4&98.4&98.6&72.7&82.3 \\
    \hline
    \end{tabular}
    }
    \caption{Evaluation results on monotonicity. ``Prec'', ``Rec'', ``F'' indicate precision, recall, and F-score.}
    \label{tab:allquantmonoall}
\end{table*}

\section{Evaluation on systematicity of 
quantifiers and negation}
We also analyze whether models can generalize to unseen combinations of quantifiers and negation.
Here, we generate Basic set~1 by setting an arbitrary quantifier to a primitive quantifier and combining it with negation.
As in (\ref{ex:1b}), we fix the primitive quantifier to the existential quantifier \textit{one} and generate the negated sentence \textit{One tiger did not run}.
Next, as in (\ref{ex:2a}) and (\ref{ex:2b}), we generate Basic set~2 by combining a primitive term (e.g., \textit{tiger}) with various quantifiers.
If a model has the ability to systematically understand primitive combinations in Basic set,
it can represent a new meaning representation with different combinations of quantifiers and negations, like (\ref{ex:3a}) and (\ref{ex:3b}).

\begin{exe}
\ex 
\begin{xlist}
\ex \label{ex:1a} \textbf{One} tiger ran
\ex \label{ex:1b} \textbf{One} tiger \textit{did not} run
\end{xlist}
\ex 
\begin{xlist}
\ex \label{ex:2a} \textbf{Every} tiger ran
\ex \label{ex:2b} \textbf{Two} tigers ran
\end{xlist}
\ex
\begin{xlist}
\ex \label{ex:3a} \textbf{Every} tiger \textit{did not} run
\ex \label{ex:3b} \textbf{Two} tigers \textit{did not} run
\end{xlist}
\end{exe}

Table~\ref{tab:quantnegall} shows the accuracy on combinations of quantifiers and negations by quantifier type.
Similar to the results with unseen combinations of quantifiers and modifiers, models can easily generalize to problems involving quantifiers of the same type as the primitive quantifier.
Table~\ref{tab:phenoneg} shows the accuracy on combinations of quantifiers and negations by modifier types.
Similar to the results in Table~\ref{tab:pheno50Kevery}, the accuracies on problems involving logical connectives or adverbs were slightly better than those on problems involving adjectives.

\begin{table*}[h]
    \centering
    \scalebox{0.8}{
    \begin{tabular}{l|cccc|cccc}\hline
    \multirow{2}{*}{Test}&\multicolumn{4}{c|}{GRU}&\multicolumn{4}{c}{Transformer}\\
    &FOL&DRS&DRS (cnt)&VF&FOL&DRS&DRS (cnt)&VF\\ \hline
    \multicolumn{9}{c}{primitive quantifier: existential quantifier \textit{one}}\\ \hline
    \Repl{Exi}&65.9&88.0&98.1&94.5&99.9&96.6&97.6&45.0\\
    \Repl{Num}&48.4&71.0&96.7&54.5&65.8&86.6&98.8&26.3\\ 
    \Repl{Uni}&22.3&0.0&52.0&53.1&0.0&0.0&70.2&26.6\\ \hline
    Valid&96.6&98.1&100.0&99.8&100.0&100.0&100.0&100.0\\ \hline
    \multicolumn{9}{c}{primitive quantifier: numeral \textit{two}}\\ \hline
    \Repl{Exi}&36.4&63.6&89.8&13.9&14.4&49.6&86.9&11.5\\
    \Repl{Num}&40.0&66.7&94.1&21.5&33.0&59.1&87.6&14.3\\ 
    \Repl{Uni}&15.4&0.0&50.7&0.0&0.0&0.0&71.0&12.4\\ \hline
    Valid&96.4&97.8&100.0&99.3&100.0&100.0&100.0&100.0\\ \hline
    \multicolumn{9}{c}{primitive quantifier: universal quantifier \textit{every}}\\ \hline
    \Repl{Exi}&12.8&0.0&73.7&78.6&0.8&0.5&52.2&59.4\\
    \Repl{Num}&17.1&0.0&75.3&78.1&0.0&0.6&59.2&65.1\\ 
    \Repl{Uni}&91.1&88.3&97.4&94.9&86.6&70.1&92.3&76.6\\ \hline
    Valid&98.8&98.1&100.0&98.8&100.0&100.0&100.0&100.0\\ \hline
    \end{tabular}
    }
    \caption{Accuracy on combinations of quantifiers and negation by quantifier type. ``DRS (cnt)'' columns show accuracies of predicted DRSs by \textsc{Counter}. ``Valid'' row shows the validation accuracy.}
    \label{tab:quantnegall}
\end{table*}

\begin{table*}
    \centering
    \scalebox{0.8}{
    \begin{tabular}{l|cccc|cccc}\hline
    \multirow{2}{*}{Test}&\multicolumn{4}{c|}{GRU}&\multicolumn{4}{c}{Transformer}\\
    &FOL&DRS&DRS (cnt)&VF&FOL&DRS&DRS (cnt)&VF\\ \hline
    \multicolumn{9}{c}{primitive quantifier: existential quantifier \textit{one}}\\ \hline
    \Repl{Adj}$+$\Repl{Neg}&34.6&33.2&68.3&45.8&19.4&45.1&84.1&25.3\\ 
    \Repl{Adv}$+$\Repl{Neg}&38.0&36.3&77.9&53.0&43.2&54.8&88.1&37.5\\ 
    \Repl{Con}$+$\Repl{Neg}&36.8&33.2&73.2&54.7&47.4&52.4&85.4&37.0\\ \hline
    \multicolumn{9}{c}{primitive quantifier: numeral \textit{two}}\\ \hline
    \Repl{Adj}$+$\Repl{Neg}&21.2&18.2&69.3&8.0&11.6&29.3&75.9&1.2\\ 
    \Repl{Adv}$+$\Repl{Neg}&26.5&28.5&71.4&10.3&19.3&36.5&81.8&17.8\\ 
    \Repl{Con}$+$\Repl{Neg}&21.9&28.1&68.5&9.1&11.7&34.6&78.7&16.3\\ \hline
    \multicolumn{9}{c}{primitive quantifier: universal quantifier \textit{every}}\\ \hline
    \Repl{Adj}$+$\Repl{Neg}&26.7&12.4&63.9&60.5&13.9&14.6&58.0&50.7\\ 
    \Repl{Adv}$+$\Repl{Neg}&25.9&13.2&69.2&66.1&21.9&15.2&60.9&63.3\\ 
    \Repl{Con}$+$\Repl{Neg}&28.5&18.8&71.4&65.9&20.2&14.6&59.7&62.8\\ \hline
    \end{tabular}
    }
    \caption{Accuracy on combinations of quantifiers and negation by modifier type.}
    \label{tab:phenoneg}
\end{table*}

\begin{table*}[h]
\centering
\scalebox{0.7}{
\begin{minipage}{13em}
    \begin{tabular}{l}
     \multicolumn{1}{c}{(a) Gold answer} \\
         \drgvar{b1} \texttt{IMP} \drgvar{b2} \drgvar{b4}\\
         \drgvar{b2} \texttt{REF} \drgvar{x1}\\
         \drgvar{b2} \texttt{lion} \drgvar{x1}\\
         \drgvar{b2} \texttt{NOT} \drgvar{b3}\\
         \drgvar{b3} \texttt{REF} \drgvar{x2}\\
         \drgvar{b3} \texttt{REF} \drgvar{x3}\\
         \drgvar{b3} \texttt{two} \drgvar{x2}\\
         \drgvar{b3} \texttt{bear} \drgvar{x2}\\
         \drgvar{b3} \texttt{three} \drgvar{x3}\\
         \drgvar{b3} \texttt{monkey} \drgvar{x3}\\
         \drgvar{b3} \texttt{chase} \drgvar{x3} \drgvar{x2}\\
         \drgvar{b3} \texttt{follow} \drgvar{x2} \drgvar{x1}\\
         \drgvar{b4} \texttt{NOT} \drgvar{b5}\\
         \drgvar{b5} \texttt{cry} \drgvar{x1}\\
     \end{tabular}
\end{minipage}
\begin{minipage}{13em}
    \begin{tabular}{l}
     \multicolumn{1}{c}{(b) GRU} \\
     \multicolumn{1}{c}{(F: 0.45)} \\
     \wrong{\drgvar{b1} \texttt{IMP} \drgvar{b2} \drgvar{b3}}\\
     \corr{\drgvar{b2} \texttt{REF} \drgvar{x1}}\\
     \corr{\drgvar{b2} \texttt{lion} \drgvar{x1}}\\
     \corr{\drgvar{b2} \texttt{NOT} \drgvar{b3}}\\
     \corr{\drgvar{b3} \texttt{REF} \drgvar{x2}}\\
     \corr{\drgvar{b3} \texttt{two} \drgvar{x2}}\\
     \corr{\drgvar{b3} \texttt{bear} \drgvar{x2}}\\
     \wrong{\drgvar{b3} \texttt{follow} \drgvar{x2} \drgvar{x2}}\\
     \corr{\drgvar{b3} \texttt{REF} \drgvar{x3}}\\
     \corr{\drgvar{b3} \texttt{three} \drgvar{x3}}\\
     \wrong{\drgvar{b4} \texttt{monkey} \drgvar{x3}}\\
     \wrong{\drgvar{b4} \texttt{like} \drgvar{x3} \drgvar{x2}}\\
     \wrong{\drgvar{b4} \texttt{like} \drgvar{x1} \drgvar{x2}}\\
     \end{tabular}
\end{minipage}
\begin{minipage}{13em}
    \begin{tabular}{l}
     \multicolumn{1}{c}{(c) Transformer} \\   
     \multicolumn{1}{c}{(F: 0.42)} \\
     \wrong{\drgvar{b1} \texttt{IMP} \drgvar{b2} \drgvar{b3}}\\
     \corr{\drgvar{b2} \texttt{REF} \drgvar{x1}}\\
     \corr{\drgvar{b2} \texttt{lion} \drgvar{x1}}\\
     \corr{\drgvar{b2} \texttt{NOT} \drgvar{b3}}\\
     \corr{\drgvar{b3} \texttt{REF} \drgvar{x2}}\\
     \corr{\drgvar{b3} \texttt{two} \drgvar{x2}}\\
     \wrong{\drgvar{b3} \texttt{monkey} \drgvar{x2}}\\
     \corr{\drgvar{b3} \texttt{follow} \drgvar{x2} \drgvar{x1}}\\
     \corr{\drgvar{b3} \texttt{REF} \drgvar{x3}}\\
     \wrong{\drgvar{b3} \texttt{john} \drgvar{x3}}\\
     \wrong{\drgvar{b3} \texttt{chase} \drgvar{x1} \drgvar{x3}}\\
     \end{tabular}
\end{minipage}
}%scalebox
\caption{Error analysis of DRSs
for the sentence ``all lions that did not follow two bears that chased three monkeys did not cry''.
Clauses in green are correct and those in red are incorrect.
``F'' shows F-score over matching clause.
}
\label{tab:errdrs}
\end{table*}

\section{Error analysis of predicted DRSs}
In the productivity experiments, the evaluation results using \textsc{Counter} on DRS prediction tasks are much higher than those by exact matching.
Table~\ref{tab:errdrs} shows an example of predicted DRSs for the sentence
\textit{all lions that did not follow two bears that chased three monkeys did not cry}. This sentence contains embedded clauses with depth two, having the following gold DRS:

\begin{center}
\scalebox{0.66}{
\drs{}{
 \ifdrs{$x_1$}{$\LF{lion}(x_1)$ \\ $\neg$ \drs{$x_2, x_3$}{$\LF{two}(x_2)$ \\ $\LF{bear}(x_2)$ \\ $\LF{three}(x_3)$ \\ $\LF{monkey}(x_3)$ \\ $\LF{chase}(x_2,x_3)$ \\ $\LF{follow}(x_1,x_2))$}}{}{$\neg$\drs{}{$\LF{cry}(x_1)$}}}
}%scalebox
\end{center}

\noindent Both GRU and Transformer tend to correctly predict some of the clauses for content words, implication, and negation that appear at the beginning of the input sentence, while they fail to capture long-distance dependencies between subject nouns and verbs (e.g., \textit{all lions ... did not cry}).
Also, \textsc{Counter} correctly evaluates the cases where the order of clauses is different from that of gold answers. 

\end{document}